\def\paperID{6}
\def\confName{CVinW}
\def\confYear{2025}

\newif\ifreview 
\newif\ifarxiv 
\newif\ifcamera \newcommand{\cameraready}{\cameratrue}
\newif\ifrebuttal 

\cameraready

\pdfoutput=1
\documentclass[10pt,twocolumn,letterpaper]{article}
\usepackage{cuted}

\ifreview \usepackage[review]{cvpr} \fi
\ifarxiv \usepackage[pagenumbers]{cvpr} \fi
\ifrebuttal \usepackage[rebuttal]{cvpr} \fi
\ifcamera \usepackage{cvpr} \fi

\usepackage{graphicx}	
\usepackage{amsmath}	
\usepackage{amssymb}	
\usepackage{booktabs}
\usepackage{times}
\usepackage{microtype}
\usepackage{epsfig}
\usepackage[table,xcdraw,dvipsnames]{xcolor}
\usepackage{caption}
\usepackage{float}
\usepackage{placeins}
\usepackage{color, colortbl}
\usepackage{stfloats}
\usepackage{enumitem}
\usepackage{tabularx}
\usepackage{xstring}
\usepackage{multirow}
\usepackage{xspace}
\usepackage{url}
\usepackage{subcaption}
\usepackage{xcolor}
\usepackage[hang,flushmargin]{footmisc}

\ifcamera \usepackage[accsupp]{axessibility} \fi

% ...

\definecolor{blue}{rgb}{0.92,0.96,1.0} 

\ifarxiv  \fi

\newcommand{\R}[1]{{%
    \textbf{%
        \ifstrequal{#1}{1}{\textcolor{red}{R#1}}{%
        \ifstrequal{#1}{2}{\textcolor{blue}{R#1}}{%
        \ifstrequal{#1}{3}{\textcolor{magenta}{R#1}}{%
        \ifstrequal{#1}{4}{\textcolor{teal}{R#1}}{%
                           \textcolor{cyan}{R#1}%
        }}}}%
    }%
}}  

\usepackage{xr-hyper}

\makeatletter
\newcommand*{\addFileDependency}[1]{
  \typeout{(#1)}
  \@addtofilelist{#1}
  \IfFileExists{#1}{}{\typeout{No file #1.}}
}

\makeatother

\definecolor{cvprblue}{rgb}{0.21,0.49,0.74}
\usepackage[pagebackref,breaklinks,colorlinks,citecolor=cvprblue]{hyperref}
\usepackage[capitalize]{cleveref}
\crefname{section}{Sec.}{Secs.}
\crefname{table}{Table}{Tables}
\crefname{figure}{Fig.}{Figs.}

\begin{document}
\title{Two by Two \includegraphics[height=0.04\linewidth]{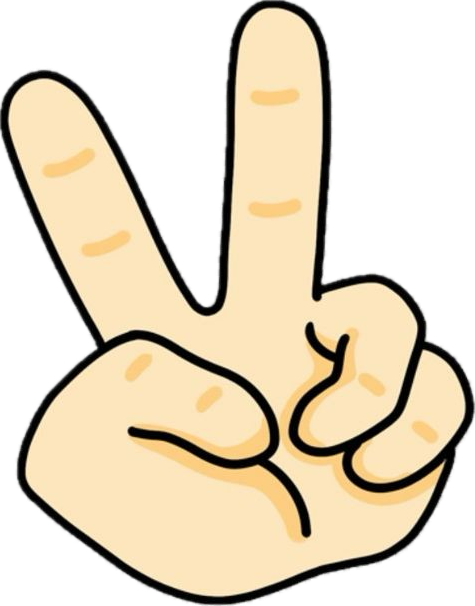}: Learning Multi-Task Pairwise Objects Assembly \\ for Generalizable Robot Manipulation}
\author{\authorBlock}
\author{
   Yu Qi \textsuperscript{2, 1*}\quad
   Yuanchen Ju\textsuperscript{1, 3*} \quad
   Tianming Wei \textsuperscript{1, 4}\quad
   Chi Chu \textsuperscript{1}\quad
   Lawson L.S. Wong \textsuperscript{2}\quad
   Huazhe Xu \textsuperscript{1, 3, 5 $\dagger$}\\
   \textsuperscript{1} Shanghai Qi Zhi Institute\quad \textsuperscript{2} Northeastern University \quad \textsuperscript{3} IIIS, Tsinghua University  \\
   \textsuperscript{4} Shanghai Jiao Tong University \quad \textsuperscript{5} Shanghai AI Laboratory 
}
\maketitle
\noindent

\begin{strip}
    \vspace{-1.6cm}
    \centering
    \includegraphics[trim=0pt 10pt 0pt 0pt, clip, width=\textwidth, height=7.5cm]{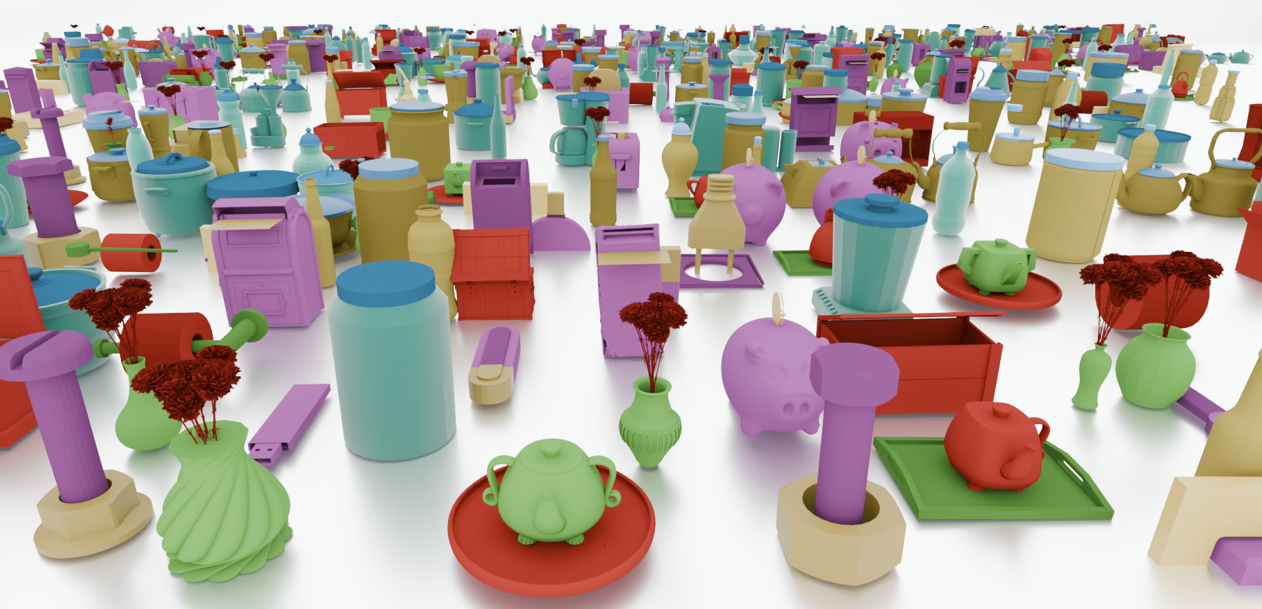}
    \captionof{figure}{\textbf{Overview of the \textit{2BY2} Dataset.} We propose the first large-scale daily pairwise object assembly dataset ~\textbf{2BY2}, which contains 1,034 instances and 517 pairwise objects with pose and symmetry annotations.}
    \label{fig:teaser}
\end{strip}

\begin{abstract}
\vspace{-0.4cm}
3D assembly tasks, such as furniture assembly and component fitting, play a crucial role in daily life and represent essential capabilities for future home robots. Existing benchmarks and datasets predominantly focus on assembling geometric fragments or factory parts, which fall short in addressing the complexities of everyday object interactions and assemblies. To bridge this gap, we present \textbf{2BY2}, a large-scale annotated dataset for daily pairwise objects assembly, covering 18 fine-grained tasks that reflect real-life scenarios, such as plugging into sockets, arranging flowers in vases, and inserting bread into toasters. 2BY2 dataset includes 1,034 instances and 517 pairwise objects with pose and symmetry annotations, requiring approaches that align geometric shapes while accounting for functional and spatial relationships between objects. Leveraging the 2BY2 dataset, we propose a two-step SE(3) pose estimation method with equivariant features for assembly constraints. Compared to previous shape assembly methods, our approach achieves state-of-the-art performance across all 18 tasks in the 2BY2 dataset.  Additionally, robot experiments further validate the reliability and generalization ability of our method for complex 3D assembly tasks. 

\end{abstract}
\vspace{-2mm}
\section{Introduction}
\label{sec:intro}
Assembly tasks are ubiquitous, such as assembling furniture, repairing household appliances, or putting together electronics. Successfully completing these tasks requires precise reasoning about the spatial relationships between pairs of objects. For robots to assist in these activities, they need to accurately estimate the 6D pose of each objects—including both their orientation and position in space. This capability is essential for domestic robots to help humans with various tasks, as it enables them to interact with their environment in a meaningful way. 
   
Daily object pairwise assembly not only requires considering the geometric constraints and spatial relationships between objects to achieve precise alignment but also needs to exhibit a certain level of generalization. Existing methods and benchmarks for solving assembly problems \cite{wang2024se, zhang20223d,liu2023puzzlenet,lu2024jigsaw,wang2024puzzlefusion++}, typically focus on matching local geometric shapes, which often results in suboptimal performance in everyday assembly scenarios which require semantic and spatial alignment. This is because they are primarily trained and tested on existing assembly datasets that consist of large-scale geometric fragments, such as Breaking Bad \cite{sellan2022breaking} and Neural Shape Mating \cite{chen2022neural}. Compared with existing assembly tasks that focus on putting together fractures of objects, daily pairwise assembly tasks are more challenging and hold greater practical significance in human life.

To bridge this gap, we introduce \textit{\textbf{2BY2}}, the first large-scale daily pairwise assembly dataset comprising 18 fine-grained tasks, shown as Figure~\ref{fig:teaser}. Compared to previous datasets and benchmarks, \textit{2BY2} contains 1,034 instances and 517 pairwise objects with pose and symmetry annotations, covering a variety of pairwise assembly tasks reflecting everyday scenarios, which require approaches that align geometric shapes while accounting for functional and spatial relationships between objects, as shown in Table~\ref{tab:dataset compare}.

\begin{table*}[!h]
    \centering
    \resizebox{1.0\textwidth}{!}{
    \begin{tabular}{l|cccccccc}
        \toprule
        \textbf{Dataset} & \textbf{\#OC} & \textbf{\#OS} & \textbf{Task Number} & \textbf{Pair} & \textbf{Task Hierarchy} & \textbf{Everyday Scenario} & \textbf{Symmetry} & \textbf{Assemble Type} \\
        \midrule
        PartNet~\cite{mo2019partnet}        & 24 & 26,671   & -    & No  & No   & No  & No & Semantic\\
        AutoMate~\cite{jones2021automate}      & 2 & 92,529   & 1     & No   & No  & No &  No & Geometric \\
        JoinABLe~\cite{willis2022joinable}      & 6  & 8,251    & 1     & No  & No & No & No  & Semantic  \\
        NSM dataset\cite{chen2022neural}   & 11  & 1,246  & 1      & Yes & No & No & No & Geometric \\
        Breaking Bad\cite{sellan2022breaking}   & - & 10,474 & 1 & No & No & No & No & Geometric \\
        Factory\cite{narang2022factory}  & 8 & 60 & 8 & Yes & No & No & No & Geometric\\
        \midrule
        \rowcolor{orange!20}
        \textbf{2BY2 Dataset (Ours)}      & \textbf{36} & \textbf{1034} & \textbf{18}   & \textbf{Yes} & \textbf{Yes} & \textbf{Yes}  & \textbf{Yes} & \textbf{Geometric and Semantic} \\
        \bottomrule
    \end{tabular}}
    \caption{\textbf{Dataset Comparison.} We compare ~\textit{2BY2} dataset with exsiting datasets and benchmarks. \textbf{\#OC} stands for the number of object categories. \textbf{\#OS} stands for the number of object shapes. \textbf{Pair} denotes whether the dataset is pairwise.    \textbf{Task Number} refers to the number of distinct assembly tasks, with the assembly of fractured pieces considered as a single task. \textbf{Task Hierarchy} stands for the different categories of task from coarse to fine, with ours shown in Section~\ref{sec:dataset overview}. \textbf{Everyday Scenario} means whether the assemble task has practical significance in real-world human applications. \textbf{Symmetry} denotes whether the dataset contains part symmetry annotation.}
    \label{tab:dataset compare}
\end{table*}

Building on this dataset, we propose a two-step pairwise network architecture for assembly tasks. Mimicking the human assembly process like we firstly put the vase on the table and then arrange flower in it, our approach predicts the pose of each object in a step-by-step manner to assemble them to a predefined canonical space, which refers to a standard coordinate system that aligns with the principles of the human world, with detailed definition in Section~\ref{sec:data annotation}. The network leverages a custom two-scale Vector Neuron DGCNN~\cite{deng2021vector} encoder with spherical convolution~\cite{cohen2018spherical} to extract SE(3) equivariant and SO(3) invariant features from point cloud inputs. Additionally, a feature fusion module and a two-step training and evaluation strategy are used to improve pose prediction accuracy.

We evaluate our approach on 18 tasks in ~\textit{2BY2} dataset to demonstrate the effectiveness on multi-task object pairwise assembly prediction. Compared to existing baselines, our method achieves an average improvement of \textit{0.046} in translation RMSE and \textit{8.97} in rotation RMSE. Moreover, we validate the effectiveness of our approach on three multi-category task, namely ~\textit{Lid Covering}, ~\textit{Inserting} and ~\textit{High Precision Placing}, as well as the ~\textit{All} task, which is defined in Section~\ref{tab:dataset overview}. Besides, real-world robot experiments validate the practical applicability of our approach.

Our main contributions are listed as follows:

1. We introduce \textbf{\textit{2BY2}}, the first large-scale daily pairwise object assembly dataset. By providing comprehensive pose and symmetry annotations for 517 pairwise objects across 18 fine-grained tasks, 2BY2 pushes the boundaries of real-world 3D assembly challenges and establishes a benchmark for pairwise assembly tasks.

2. Our two-step pairwise SE(3) pose estimation method, leveraging equivariant geometric features, demonstrates superior performance compared to existing shape assembly methods, significantly reducing translation and rotation errors and enhancing the accuracy of 6D pose estimation.

3. Our approach achieves state-of-the-art performance on the benchmark, with real-world robot experiments demonstrating its capability, providing a generalizable solution for robot manipulation using pairwise object assembly.

\section{Related Work}

\subsection{Object Assembly Benchmarks and Datasets}
Object reassembly has led to various datasets in computer vision and robotics. In computer vision, datasets like AutoMate~\cite{jones2021automate} and JoinABLe~\cite{willis2022joinable} focus on reassembling fragments using geometric clues, while early datasets~\cite{brown2008system, funkhouser2011learning, huang2006reassembling, shin2012analyzing} were limited in scale. Recent efforts, such as Neural Shape Mating~\cite{chen2022neural} and Breaking Bad~\cite{sellan2022breaking}, generate large-scale fractured object data using parametric segmentation. In robotics, benchmarks like Factory~\cite{narang2022factory}, RLBench~\cite{james2020rlbench}, and RoboSuite~\cite{zhu2020robosuite} lack diverse shapes and assembly tasks under varying initial poses. In contrast, our dataset includes over 500 diverse object pairs across 3 categories and 18 assembly tasks, providing a comprehensive benchmark for pairwise object assembly, supporting the development of generalizable methods for real-world applications.

\subsection{3D Shape Assembly}
3D shape assembly~\cite{funkhouser2004modeling, litvak2019learning, luo2019reinforcement, xiong2023learning, zakka2020form2fit}, also known as part assembly, involves reconstructing objects from fragments, such as shattered sculptures or disassembled furniture. Existing methods use graphical models~\cite{chaudhuri2011probabilistic, jaiswal2016assembly, kalogerakis2012probabilistic} and neural networks~\cite{li2020learning, xu2023unsupervised, Cheng2023ScorePAS3, wang2022ikea, xu2024spaformer, zhang2024scalable, wu2020pq, yin2020coalesce, ju2025robo, zhu2024densematcher} to capture geometric and semantic relationships. Approaches such as~\cite{chen2022neural, narayan2022rgl, zhang20223d, zhan2020generative} focus on pose estimation and part assembly without relying on predefined semantic information. Few-shot learning has been applied to assembly tasks~\cite{li2023rearrangement}, while jigsaw puzzle techniques~\cite{noroozi2016unsupervised, lu2024jigsaw} leverage shape completion strategies. Recent works~\cite{wang2024puzzlefusion++, hossieni2024puzzlefusion, scarpellini2024diffassemble, huang2024imagination} utilize diffusion models to refine poses or point clouds for assembly. In contrast, our method introduces a two-step pairwise network for step-by-step assembly, tailored to pairwise object alignment.

\subsection{6D Pose Estimation for Robot Manipulation}
6D pose estimation is crucial in robotics and computer vision for object interaction in unstructured environments \cite{xiang2018posecnn, hinterstoisser2012, Tremblay2018DeepOP}. Early handcrafted feature-based methods struggled in cluttered scenes \cite{hinterstoisser2011, hinterstoisser2012}, while CNN-based approaches improved performance but lacked generalization \cite{xiang2018posecnn, wadim2017}. Domain randomization enhances robustness by varying synthetic datasets \cite{Tremblay2018DeepOP, joshua2017}. In assembly tasks, 6D pose estimation aids manipulation planning with predefined objects \cite{www2016, ismael2019}. Like \cite{huorbitgrasp, huang2024imagination, huang2024match, huang2023edge, zhu2023robot, xue2023useek, gao2024riemann, tie2025etseed, zhao2025hierarchical, zhu2025equact, zhu2026residual, zhao2026generalizable, qi2025bear, huang2026pix2act}, our method leverages equivariant features for efficient 6D pose learning and improved generalization.

\section{2BY2 Dataset}
\label{sec:dataset}

\subsection{2BY2 Dataset Overview}
\label{sec:dataset overview}

We present the first large-scale 3D pairwise object assembly dataset for everyday scenarios, with detailed annotations for each object pair. The meshes in our dataset come from 3D Warehouse~\cite{3dwarehouse}, SAPIEN PartNet-Mobility~\cite{Xiang_2020_SAPIEN}, Google SketchUp 3D Challenge~\cite{sketchupchallenge}, and Objaverse~\cite{deitke2024objaverse}. These meshes are manually paired, cleaned, annotated, and scaled uniformly. The 2BY2 dataset contains 517 unique pairs across three main tasks: ~\textit{Lid Covering}, ~\textit{Inserting}, and \textit{High Precision Placing}, further subdivided into multiple subcategories, as shown in Table~\ref{tab:dataset overview}.

\begin{table*}[h]
    \centering
    \resizebox{1.0\textwidth}{!}{
    \begin{tabular}{c | c c c c c | c c c c c c c c | c c c c c c}
        \toprule
        \raisebox{-1.5ex}{\textbf{Task}} 
        & \multicolumn{5}{c}{\textbf{Lid Covering}} 
        & \multicolumn{8}{c}{\textbf{Inserting}} 
        & \multicolumn{5}{c}{\textbf{High Precision Placing}} \\  
        \cmidrule(lr){2-19}
        \textbf{} 
        & \textbf{Kit} & \textbf{Bot} & \textbf{Ket} & \textbf{Cof} & \textbf{Cup} 
        & \textbf{Plu} & \textbf{Chi} & \textbf{Let} & \textbf{Bre} & \textbf{Nut} 
        & \textbf{Coi} & \textbf{Key} & \textbf{Usb} 
        & \textbf{Box} & \textbf{Tis} & \textbf{Flo} & \textbf{Tea} & \textbf{Pos} \\
        \midrule
        \textbf{Pair Num} 
        & 24 & 86 & 28 & 26 & 16 
        & 14 & 19 & 32 & 24 & 20 
        & 21 & 20 & 20 
        & 25 & 20 & 60 & 42 & 21 \\
        \bottomrule
    \end{tabular}
    }
    \caption{\textbf{2BY2 Dataset Statistics Overview.} The figure presents the number of object pairs across all task categories in the 2BY2 dataset, where each pair consists of two unique objects. The first row categorizes tasks into three major groups, while the second row provides a detailed breakdown of specific task categories. Specifically, \textbf{Kit} = Kitchenport, \textbf{Bot} = Bottle, \textbf{Ket} = Kettle, \textbf{Cof} = Coffee machine, \textbf{Cup} = Cup, \textbf{Plu} = Plug into socket, \textbf{Chi} = Children's toy, \textbf{Let} = Letter into mailbox, \textbf{Bre} = Bread into toaster, \textbf{Nut} = Bolt into nut, \textbf{Coi} = Coin into piggy bank, \textbf{Key} = Key into lock, \textbf{Usb} = USB cap, \textbf{Box} = Shoe boxing, \textbf{Tis} = Tissue placement on rack, \textbf{Flo} = Flower into vase, \textbf{Tea} = Teaware arrangement on tray, \textbf{Pos} = Positioning a cup on the coffee machine for coffee dispensing.}
    \label{tab:dataset overview}
\end{table*}

\subsection{Data Annotation}
\label{sec:data annotation}
To ensure high quality and reliability of our dataset, we conducted systematic cleaning and annotation of the collected meshes. First, we manually segment, integrate, and pair the meshes, classify them into ~\textit{Object B} and ~\textit{Object A}. ~\textit{Object B} is the base or the receiving component, such as the nut, the vase, the postbox. \textit{Object A}, is the fitting component, such as the bolt, the flower, the mail.  This classification aligns with intuitive human assembly logic and supports our network’s prediction strategy, such as positioning a nut before the bolt, as detailed in Sections~\ref{sec:problem formulation} and ~\ref{sec:method overview}. Automated scripts were used to uniformly scale meshes and align each pair to a canonical pose in world frame, defined as the object resting stably on the \textit{XY} plane with its lowest point aligned to \textit{Z=0}. For instance, bottles and vases are aligned as if placed on a table, and mailboxes on the ground.

During point cloud generation, we utilized blue noise sampling method~\cite{sellan2022breaking} to extract point clouds uniformly from each mesh surface with dimension $(1024,3)$. We also annotated each object category with its inherent symmetry properties, specifically considering rotational symmetry along the \textit{Z}-axis, such as bottles, screws, and mirror symmetry along the \textit{X}-axis, such as bread, letters.

\subsection{Data Division and Task Diversity Analysis}

\begin{figure}[h!]
    \centering
    \includegraphics[width=\linewidth,  height=4.4cm, trim={15, 27, 23, 20}, clip]{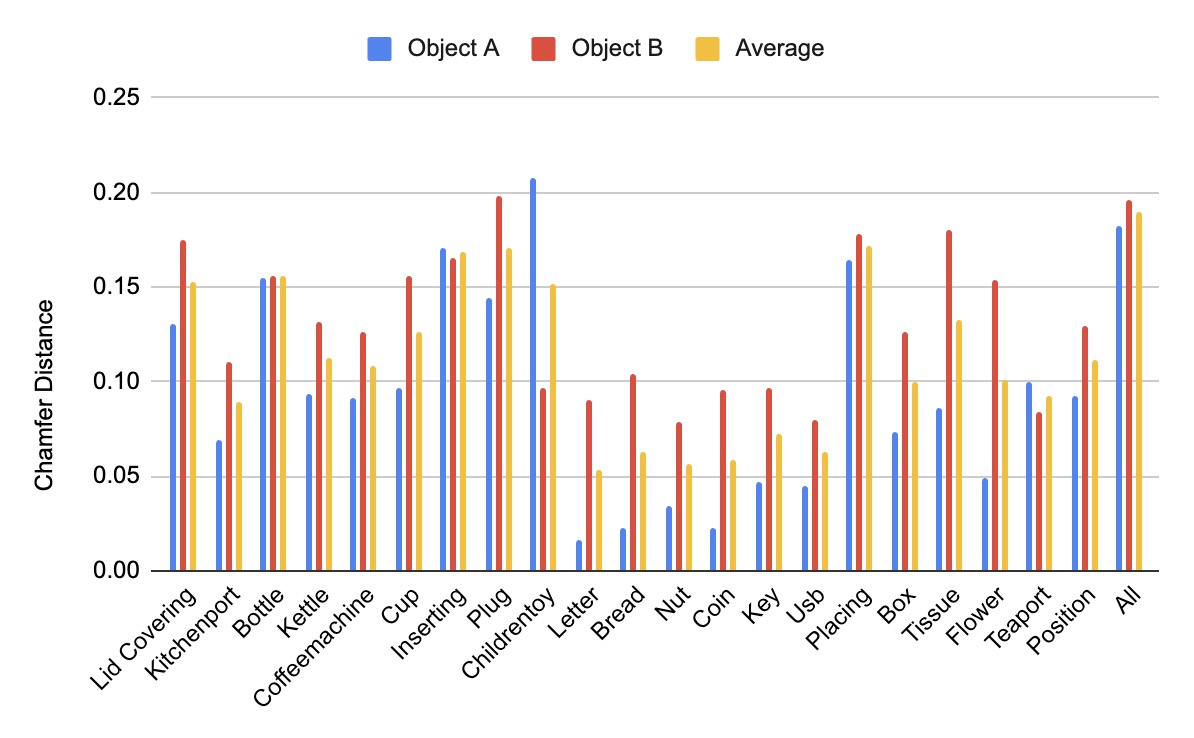}
    \caption{\textbf{Chamfer Distance Between Training and Testing Set.} We normalize point clouds and compute the Chamfer Distance. For each task we calculate the distance separately between point cloud of ~\textit{Object A} and ~\textit{Object B} in the training set and test set.} 
    \label{fig:chamfer distance}
\end{figure}

\begin{figure}[h!]
    \centering
    \includegraphics[width=\linewidth, height=5.3cm, trim={0, 0, 0, 0}, clip]{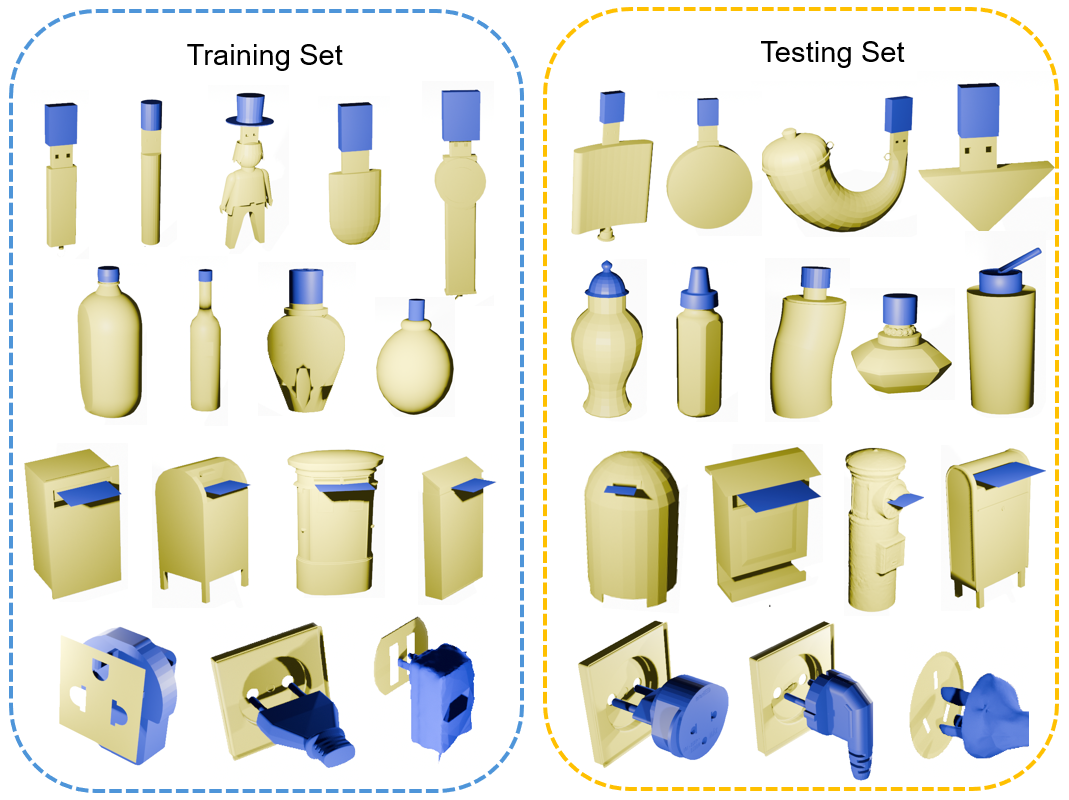}
    \caption{\textbf{Task Diversity Visualization.} The image shows selected objects from four different tasks: \textit{USB}, \textit{Bottle}, \textit{Letter}, and \textit{Plug in Socket}. On the left are the objects selected on training set, and on the right is the testing set. As seen in the legend, object geometry varies in both the training and testing set, with the testing set containing novel shapes not seen in the training set.} 
    \label{fig:task diversity}
\end{figure}

Our dataset provides diverse task coverage across categories, with each further divided into specific sub-categories, see Table~\ref{tab:dataset overview}. Objects within each category vary in shape, size, and type. To enhance generalization, the testing set includes objects with unseen geometric shapes, as shown in Figure~\ref{fig:task diversity}. We also compute Chamfer Distance on point clouds between training and testing sets to quantify geometry differences, as shown in Figure~\ref{fig:chamfer distance}. This diversity ensures generalization ability and applicability in real world scenarios and supports robust 3D matching and assembly tasks.

\vspace{-0.5cm}
\section{Problem Formulation}
\label{sec:problem formulation}
The task takes two point clouds as input, namely $\mathcal{P}_A$ and $\mathcal{P}_B$, each with dimension $(1024, 3)$. These point clouds are derived from objects $\mathcal{O}_A$ and $\mathcal{O}_B$ from predefined canonical pose, as detailed in ~\ref{sec:data annotation}, respectively, and is randomly augmented with SO(3) rotation and being translated to its centroid. The desired output would be two individual SE(3) pose two assemble $\mathcal{O}_A$ and $\mathcal{O}_B$ to the canonical pose. 

\section{Method}
\label{sec:method}

\subsection{Two-step Pairwise Network Architecture}
\label{sec:method overview}
\begin{figure*}[h!]
    \centering
    \includegraphics[width=\linewidth, trim={0, 0, 0, 0}, clip]{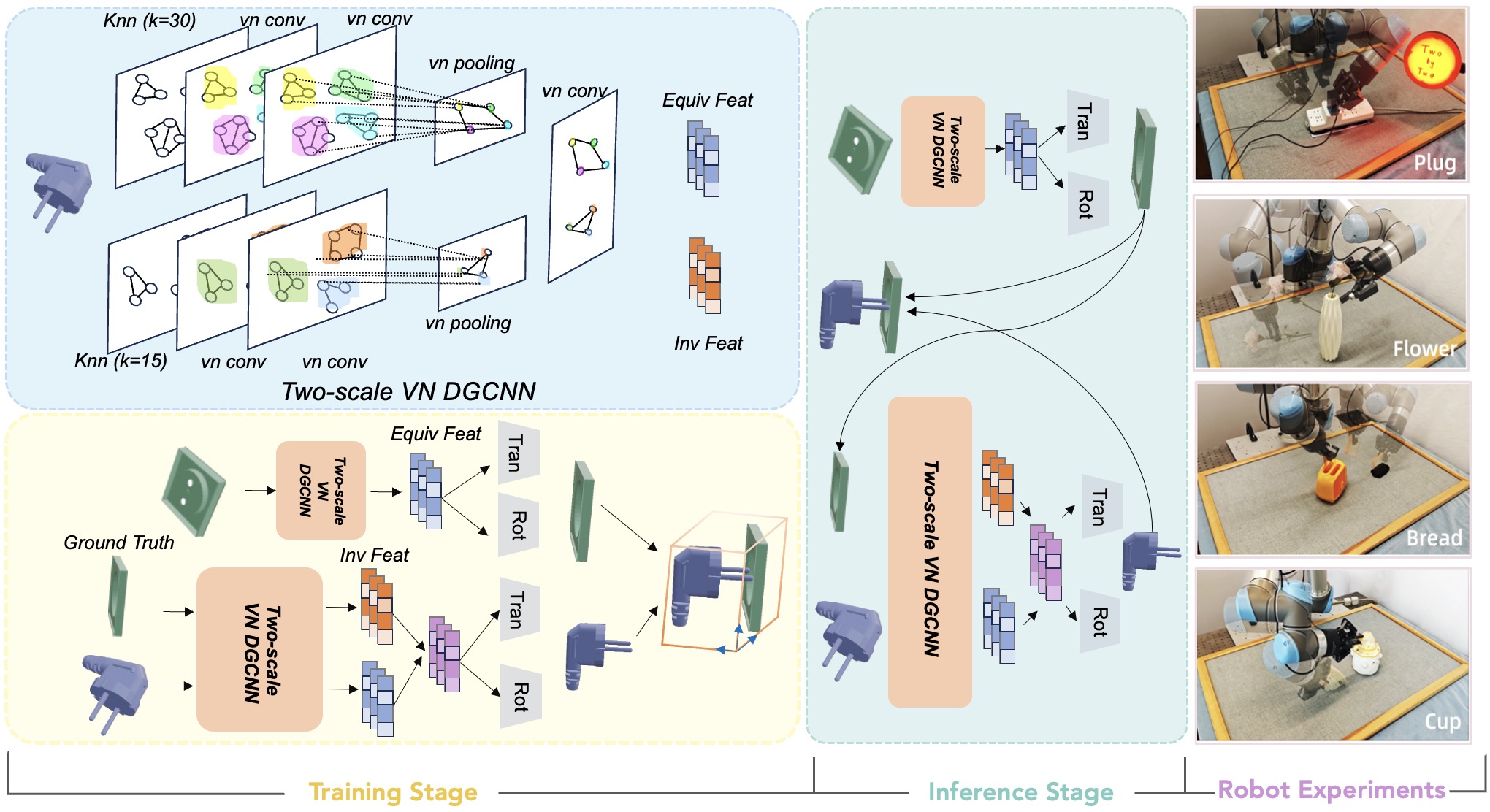}
    \caption{\textbf{Our Two-Step Pairwise Network.} We utilize two-scale VN DGCNN as our encoder to extract equivariant and invariant feature. We first predict the canonical pose of $\mathcal{O_B}$ and then predict the pose of $\mathcal{O_A}$ according to it.
    }
    \label{fig:pipeline}
\end{figure*}

To effectively learn pairwise object assembly, we propose a two-step pairwise network architecture with two branches: Branch B ($\mathcal{B_B}$) and Branch A ($\mathcal{B_A}$), as shown in Figure~\ref{fig:pipeline}. Branch B predicts the pose of $\mathcal{P_B}$, which is the socket, using a two-scale Vector Neuron DGCNN encoder~\cite{deng2021vector} to extract SE(3) equivariant features, denoted as $\mathcal{E_B}$, followed by MLP-based pose prediction heads for translation and rotation. The transformed $\mathcal{P_B}$ and inserter object $\mathcal{P_A}$, which is the plug, are then passed to Branch A, which extracts SE(3) equivariant features, denoted as $\mathcal{E_A}$ and SO(3) invariant features ($\mathcal{I_B}$). The features are fused through element-wise multiplication, allowing $\mathcal{B_A}$ to predict the pose of $\mathcal{P_A}$ using information from both objects. This architecture ensures geometric alignment and matching by leveraging shared feature representations while reducing feature interference.

Our two-step pairwise network is inspired by the human approach to pairwise assembly tasks. For example, when arranging a vase with flowers, one intuitively first positions the vase correctly before placing the flowers inside. Similarly, inserting an envelope into a mailbox requires identifying the mailbox slot’s pose first. By mimicking this sequential strategy, our model simulates human decision-making process, enabling more efficient and accurate assembly tasks.

\subsection{Two-scale SE(3) Equivariant and SO(3) Invariant Feature Extraction}
\label{sec: exp encoder}

We employ a two-scale SE(3) Vector Neuron DGCNN, an enhanced variant of the original Vector Neuron DGCNN~\cite{deng2021vector}, as our encoder to extract SE(3) equivariant and SO(3) invariant features. This architecture leverages equivariance to improve sample efficiency of the model, while incorporating a two-scale information fusion mechanism to capture geometric features at two different scales.

\textbf{SE(3) Equivariance and SO(3) Invariance.} SE(3) equivariance combines SO(3) rotation and T(3) translation equivariance: rotation equivariance ensures that a network’s output rotates with the input, while translation equivariance shifts the output accordingly. SO(3) invariance means the network’s output remains unchanged under any 3D rotation. By leveraging SE(3) equivariance, the model benefits from improved sample efficiency and generalization. This is particularly advantageous for assembly tasks, where objects may appear in arbitrary poses.

\textbf{Vector Neuron DGCNN.} The Vector Neuron Network~\cite{deng2021vector} extends traditional neurons from scalars to 3D vectors, designs vector-based convolutional layers and non-linear functions like pooling and ReLU to support SO(3) equivariant and SO(3) invariant feature extraction. VNN operates in vector space, captures richer geometric relationships and ensures more robust feature representations for downstream tasks. 

\textbf{Two-scale Vector Neuron DGCNN.} We propose a two-scale Vector Neuron DGCNN for extracting SE(3) equivariant and SO(3) invariant features $\mathcal{E_B}$, $\mathcal{I_B}$, and $\mathcal{E_A}$. As shown in Figure~\ref{fig:pipeline}, the encoder comprises two branches with different $K$ values, each consisting of multiple Vector Neuron convolutional layers followed by pooling. The extracted features from both branches are concatenated and further processed through an additional Vector Neuron convolutional layer. Point clouds $\mathcal{P_B}$ and $\mathcal{P_A}$ are independently processed, forming graphs that propagate through both branches.

The SO(3) rotation equivariance of our encoder is ensured by the inherent equivariant properties of the Vector Neuron layers. To achieve T(3) translation equivariance, with an input point cloud $P=(p_1, p_2, ..., p_n), p_i\in R^3$, we compute its centroid $x = (\Sigma_{i=1}^{n}p_i)/n$, and get the input point cloud as $P' = P - x$. In this way, our prediction is T(3) translation equivariant, i.e., $f$ is our encoder and $\mathcal{P}$ is the original point cloud.
\begin{equation}
f(\mathcal{P} + \mathcal{T}) = f(\mathcal{P}) + \mathcal{T}, \quad T \in R^3 
\end{equation}

Our two-scale VN DGCNN employs dual K-nearest neighbor (KNN) values to extract features across two distinct scales, enhancing its ability to capture both local and global information. This pyramid structure enables the network to simultaneously grasp overall object shapes and fine-grained details, improving feature extraction.

\subsection{Cross Object Fusion Module}

We utilize point-wise multiplication, shown in Figure~\ref{fig:pipeline}, as our cross object fusion module designed in $B_A$. We fuse the feature of $\mathcal{P_B}$ and $\mathcal{P_A}$ by multiplying $\mathcal{I_B}$ and $\mathcal{E_A}$, so that each point in $\mathcal{P}_A$ will have the geometry feature of both $\mathcal{P}_A$ and $\mathcal{P}_B$. This approach integrates the geometric feature of $\mathcal{P}_B$ in each point while preserving the rotation equivariance of $\mathcal{P}_A$, i.e., $f$ is an equivariant neural network, $R$ is random rotation matrix,
\begin{equation}
f(R\cdot(\mathcal{I_B} * \mathcal{E_A})) = R\cdot f(\mathcal{I_B} * \mathcal{E_A}), \quad R \in R^{3\times3}
\end{equation}

\subsection{Pose Prediction}

At both branches, we utilizes two seperate MLPs as our pose prediction head, to separately predict the translation $T \in R^3$ and rotation $R \in R^{3\times 3}$. 
Compared to predicting translation and rotation within a single prediction head, this approach helps mitigate the issue of differing convergence speeds between the two components.

\subsection{Training and Evaluation Strategy}

We adopt a separate training and evaluation strategy for our network. To minimize the impact of pose prediction errors of $\mathcal{P_B}$ on  $\mathcal{P_A}$, we train $\mathcal{B_A}$ and $\mathcal{B_B}$ independently. Specifically, for $\mathcal{B_A}$, during training, we utilize $\mathcal{P_B}$ under canonical pose, which is our ground truth point cloud of $\mathcal{P_B}$, to train our model. During testing, we first predict the pose of $\mathcal{P_B}$, then use the transformed $\mathcal{P_B}$, along with the initial $\mathcal{P_A}$ to predict A's pose, as shown in Figure~\ref{fig:pipeline}. This phased, two-step training and evaluation strategy reduces errors caused by joint training of object poses, ensuring more accurate predictions.

\subsection{Loss Function}

To train our network to robustly predict poses, we use the following equation as our loss function:
\begin{equation}
\mathcal{L} = \lambda_{\text{rot}} \mathcal{L}_{\text{rot}} + \lambda_{\text{trans}} \mathcal{L}_{\text{trans}}
\end{equation}

Specifically, for predicted pose translation $T_{\text{pred}} \in R^3 $, rotation $\mathcal{R}_{\text{pred}} \in R^{3\times 3}$ and ground truth pose translation $T_{\text{gt}} \in R^3$ and rotation $\mathcal{R}_{\text{gt}} \in R^{3\times 3}$, we use $\mathcal{L}_1$ loss to compute our $\mathcal{L}_{\text{trans}}$:
\begin{equation}
\mathcal{L}_{\text{trans}} =  \mathcal{L}_1(T_{\text{pred}}, T_{\text{gt}})
\end{equation}

As for the rotation, we utilize Geodesic Distance, which measures the shortest path between two rotations on the rotation manifold.  It offers a smooth and bounded angular error, ensuring stable gradients and accurately achieving precise rotation alignment.
\begin{equation}
\mathcal{L}_{\text{rot}} = \arccos\left(\frac{\text{tr}(\mathcal{R}_{\text{gt}} \mathcal{R}_{\text{pred}}^{T}) - 1}{2}\right)
\end{equation}

\section{Experiments}
\label{sec:experiment}

\begin{table*}[h]
\centering
\resizebox{1.0\textwidth}{!}{%
\begin{tabular}{l|cc|cc|cc|cc|cc}
\toprule
\textbf{Task} & \multicolumn{2}{c}{\textbf{Jigsaw~\cite{lu2024jigsaw}}} & \multicolumn{2}{c}{\textbf{Puzzlefusion++~\cite{wang2024puzzlefusion++}}} & \multicolumn{2}{c}{\textbf{NSM~\cite{chen2022neural}}} & \multicolumn{2}{c}{\textbf{SE(3)-Assembly~\cite{wu2023leveraging}}} & \multicolumn{2}{c}{\textbf{Ours}} \\
& RMSE(T) & RMSE(R) & RMSE(T) & RMSE(R) & RMSE(T) & RMSE(R) & RMSE(T) & RMSE(R) & RMSE(T)$\downarrow$ & RMSE(R)$\downarrow$ \\
\midrule
\textbf{Lid Covering}  & 0.398 & 33.33 & 0.408 & 37.74 & 0.184 & 33.45 & 0.125 & 21.37 & \textbf{0.090} & \textbf{16.12} \\
Kitchenport  & 0.477 & 45.80 & 0.423 & 47.23 & 0.237 & 57.47 & 0.093 & 17.29 & \textbf{0.068} & \textbf{16.60} \\
Bottle       & 0.411 & 34.71 & 0.385 & 35.23  & 0.227  & 78.58 & 0.147  & 36.92 & \textbf{0.076} & \textbf{27.70} \\
Kettle        & 0.335 & 43.71 & 0.372 & 38.38 & 0.215 & 61.10  & 0.133 & 13.15 & \textbf{0.111} & \textbf{11.56} \\
Coffeemachine   & 0.527 & 32.67 & 0.437 & 34.64  & 0.253 & 50.66  & 0.142 & 24.43  & \textbf{0.076} & \textbf{22.83} \\
Cup    & 0.408 & 33.55 & 0.439 & 33.58 & 0.260 & 67.35  & 0.160 & 46.46 & \textbf{0.122} & \textbf{23.18} \\
\textbf{Inserting}    & 0.364 & 53.58 & 0.327 & 57.83 & 0.275 & 69.93 & 0.199 & 46.3 & \textbf{0.142} & \textbf{38.03} \\
Plug               & 0.372 & 56.89 & 0.348 & 48.89  & 0.303 & 52.26  & 0.176 & 18.58  & \textbf{0.094} & \textbf{9.74} \\
Childrentoy                 & 0.268 & 59.88 & 0.245 & 63.21  & 0.271 & 93.77  & 0.302 & 80.52 & \textbf{0.242} & \textbf{57.81} \\
Letter         & 0.409 & 67.99 & 0.357 & 72.08  & 0.317 & 76.48  & 0.121 & 39.24  & \textbf{0.094} & \textbf{33.74} \\
Bread & 0.220 & 57.84 & 0.201 & 60.92  & 0.171 & 65.50  & 0.111 & 51.13  & \textbf{0.090} & \textbf{36.40} \\
Nut & 0.476 & 40.08 & 0.323 & 47.29  & 0.271 & 55.32  & 0.102 & 46.68  & \textbf{0.051} & \textbf{35.60} \\
Coin & 0.406 & 39.62 & 0.348 & 51.40  & 0.289 & 62.58  & 0.111 & 28.69  & \textbf{0.107} & \textbf{22.88} \\
Key & 0.384 & 42.85 & 0.348 & 50.38  & 0.290 & 63.60  & 0.087 & 17.28  & \textbf{0.045} & \textbf{16.32} \\
Usb & 0.463 & 67.41 & 0.342 & 58.23  & 0.252 & 69.90  & 0.215 & 32.28  & \textbf{0.128} & \textbf{28.98} \\
\textbf{Precision Placing} & 0.375 & 73.94 & 0.287 & 67.81  & 0.211 & 85.02  & 0.134 & 57.86  & \textbf{0.115} & \textbf{44.84} \\
Box  & 0.137 & 33.72 & 0.134  & 40.47 & 0.130  & 72.47 & 0.071  & 25.08 & \textbf{0.066} & \textbf{21.53} \\
Tissue & 0.292 & 82.39 & 0.265 & 85.18  & 0.175 & 79.78  & 0.183 & 73.02  & \textbf{0.115} & \textbf{64.37} \\
Flower & 0.328 & 64.39 & 0.283 & 59.04  & 0.246 & 87.18  & 0.213 & 64.89  & \textbf{0.125} & \textbf{42.23} \\
Teaport & 0.302 & 68.33 & 0.324 & 61.01  & 0.288 & 56.32  & 0.085 & 40.59  & \textbf{0.050} & \textbf{26.11} \\
Position & 0.423 & 58.07 & 0.389 & 57.55  & 0.257 & 70.11  & 0.166 & 28.17  & \textbf{0.141} & \textbf{24.46} \\
\midrule
\textbf{ALL}            & 0.360 & 53.34 & 0.342 & 58.23  & 0.284 & 70.30  & 0.233 & 52.34  & \textbf{0.110} & \textbf{41.44} \\
\bottomrule
\end{tabular}%
}
\caption{\textbf{Quantitative Evaluation on \textit{2BY2} for Pairwise Object Assembly.} Our method outperforms the baseline across all 18 fine-grained assembly tasks, as well as demonstrating significant improvement on three cross-category assembly tasks. It achieves a lower task average with a reduction of \textit{0.046} in translation RMSE and \textit{8.97} in rotation RMSE.}
\label{tab:pose_evaluation}
\end{table*}

In this section, we present a comprehensive evaluation and analysis of our two-step pairwise network architecture by addressing the following key questions: 

\textbf{1.} How does our network perform on 2BY2 tasks compared to existing baseline approaches, including matching-based, graph-network-based, and diffusion-based assembly methods? 

\textbf{2.} How well does our network generalize across multiple tasks within the 2BY2 dataset? Can our network effectively handle a diverse set of tasks simultaneously? 

\textbf{3.} Can our network generalize to real-world robot tasks?

\subsection{2BY2 Dataset Main Experiment}
\subsubsection{Experiment Set Up}

\textbf{Tasks.} We divide the 18 assembly tasks in the 2BY2 dataset into training and testing sets individually and compared the performance of our method with various baseline approaches. To further evaluate its cross-task generalization ability, we conducted additional experiments on tasks such as \textit{Lid Covering}, \textit{Insertion}, and \textit{High Precision Placement}, as well as \textit{All} task, which requires the method to handle all tasks in the entire dataset. See Table~\ref{tab:dataset overview} for task details.

\textbf{Evaluation metrics.} Following metrics from datasets like Breaking Bad~\cite{sellan2022breaking} and Neural Shape Mating~\cite{chen2022neural}, we use Root Mean Squared Error (RMSE) to evaluate both rotation and translation of the predicted SE(3) pose. Specifically, rotations are represented using Euler angles with symmetry considerations, see Section~\ref{sec:data annotation} for symmetry details.

\textbf{Training parameters.} We set batch size to be 4, and the initial learning rate of Adam Optimizer~\cite{diederik2014adam} to be 1e-4. We train models for 1000 epochs for them to fully converge.

\subsubsection{Baselines} We compare our method with SE-3 assembly~\cite{wu2023leveraging}, Puzzlefusion++~\cite{wang2024puzzlefusion++}, Jigsaw~\cite{lu2024jigsaw} and Neural Shape Mating~\cite{chen2022neural}. 
\begin{itemize}
    \item \textbf{SE-3 Assembly~\cite{wu2023leveraging}} proposes a network architecture to leverage SE(3) equivariance for representations considering multi-part correlations, and predict the pose of each part jointly.

    \item \textbf{Puzzlefusion++~\cite{wang2024puzzlefusion++}} proposes an auto-agglomerative 3D fracture assembly framework. It iteratively aligns and merges fragments using a diffusion model for 6-DoF alignment and a transformer model for verification.

    \item \textbf{Jigsaw~\cite{lu2024jigsaw}} leverages hierarchical
    features of global and local geometry to match and align the fracture surfaces, and recovers the global pose of
    each piece to restore the underlying object.

    \item \textbf{Neural Shape Mating~\cite{chen2022neural}}  utilizes PointNet for feature encoding and a transformer for feature fusion to learn the correlations between assembly parts, enabling joint prediction of their poses. 

\end{itemize}

\subsubsection{2BY2 Benchmark Results and Analysis}

Table~\ref{tab:pose_evaluation} presents the quantitative performance of our method compared to all baselines. The results show that our approach outperforms the baselines across 18 fine-grained assembly tasks, with an average improvement of \textit{0.046} in translation RMSE and \textit{8.97} in rotation RMSE.

Additionally, we evaluate our method on three cross-category tasks defined in Section~\ref{sec:dataset overview}, namely \textit{Lid Covering}, \textit{Inserting} and \textit{High Precision Placing}, and achieve the state-of-the-art performance. Moreover, in the most comprehensive ~\textit{All} task, we outperform the baseline by \textit{0.123} in translation and \textit{10.90} in rotation, demonstrating strong generalization across tasks and object shapes. In the meantime, baseline comparisons confirm the rigor and challenge of our tasks. Results on challenging tasks like ~\textit{Plug} and ~\textit{Key} highlight our framework's effectiveness in complex scenarios.

We analyze that the superior performance of our designed network is due to the approach of separately predicting the poses of the two objects in a step-by-step manner. This prevents the pose errors from interfering with each other, which often occurs in other baselines when predicting both poses simultaneously. Additionally, the design of our encoder makes our network more sensitive to subtle changes in rotation and translation, resulting better performance.

\begin{figure}[h!]
    \centering
    \includegraphics[width=\linewidth, trim={15, 30, 23, 20}, clip]{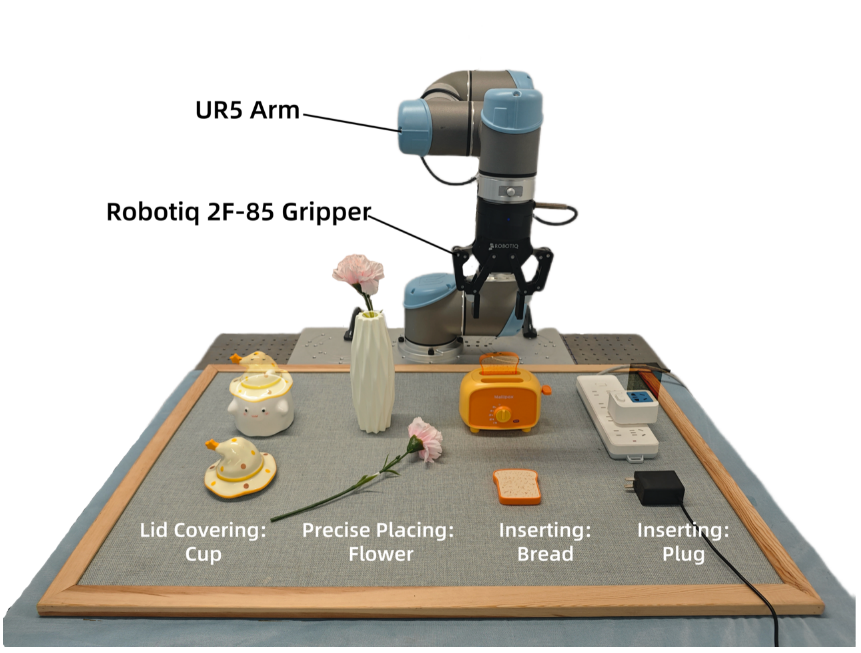}
    \caption{\textbf{Real Robot Setup.} We conduct real-world robot experiments on \textit{Cup}, \textit{Flower}, \textit{Bread} and \textit{Plug} tasks.} 
    \label{fig:realworld}
\end{figure}

\subsection{Real-World Robot Experiment}
\textbf{Real-world robot experiment setup.} As shown in Figure~\ref{fig:realworld}, we conduct our real-world robot experiments using a UR5 robotic arm, equipped with a Robotiq 2F-85 Gripper. We select four tasks, \textit{Cup}, \textit{Flower}, \textit{Bread} and \textit{Plug}, demonstrating that our model exhibits strong generalization ability on unseen real-world objects.

We place objects in the scene with random initial poses and scan them to obtain their point clouds. Using pre-trained models on selected data of \textit{2BY2} dataset, we predict the pose of each object. A manually designed grasping pose is then applied to pick up each object, and based on the predicted poses, the robotic arm plans a trajectory to complete the assembly. We use SE(3) assembly~\cite{wu2023leveraging} as the baseline and test our approach on 10 different initial poses. As shown in Table~\ref{tab:real_world exp}, our method significantly outperforms the baseline.

\begin{table}[h]
\centering
\resizebox{0.4\textwidth}{!}
{%
\begin{tabular}{lccccc}
\toprule
\textbf{Task} &\textbf{Cup}& \textbf{Flower} & {\textbf{Bread}} & \textbf{Plug} & \textbf{Overall} \\
\midrule
\textbf{SE(3)~\cite{wu2023leveraging}} & $2/10$ &$4/10$ & $1/10$ & $2/10$ & $22.5\%$  \\
\textbf{Ours} &$8/10$ & $10/10$ & $6/10$ & $7/10$ & $77.5\%$ \\
\bottomrule
\end{tabular}
}
\caption{\textbf{Real-World Robot Experiment Success Rate Results.}}
\label{tab:real_world exp}
\end{table}

\begin{figure*}[t]
    \centering
    \includegraphics[width=1\linewidth,trim={0 0 0 10}, clip]{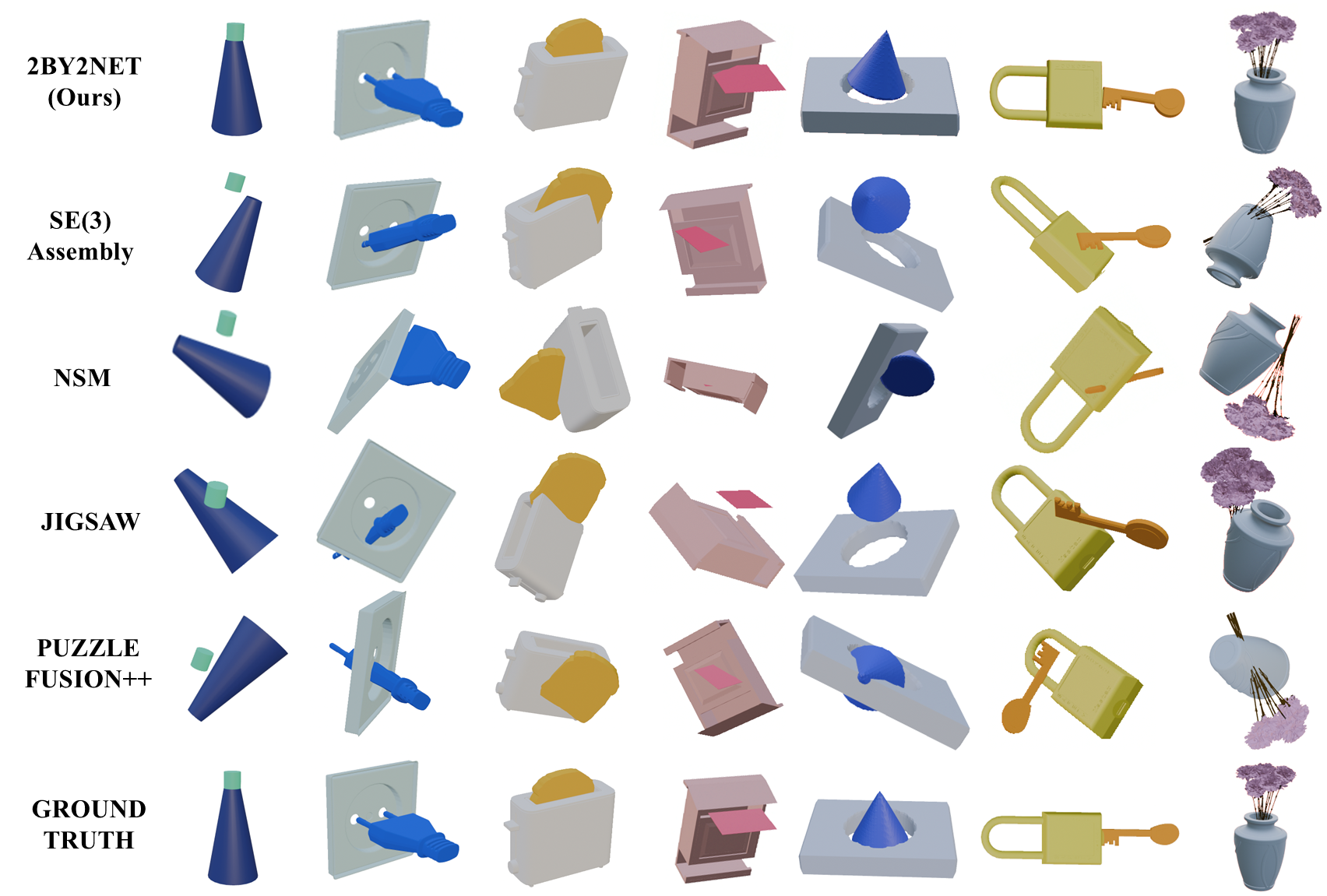}
    \caption{\textbf{Qualitative Results Comparison.} We highlight ~\textit{Bottle}, ~\textit{Plug}, ~\textit{Bread}, ~\textit{Letter}, ~\textit{Childrentoy}, ~\textit{Key}, and ~\textit{Flower} tasks to demonstrate our improved translation and rotation predictions compared to baseline methods.}
    \label{fig:evaluation}
\end{figure*}
\begin{table*}[h]
\centering
\resizebox{1.0\textwidth}{!}{%
\begin{tabular}{l|cc|cc|cc|cc|cc}
\toprule
\textbf{Task} & \multicolumn{2}{c}{\textbf{Vector Neuron DGCNN~\cite{deng2021vector}}} & \multicolumn{2}{c}{\textbf{DGCNN~\cite{wang2019dynamic}}} & \multicolumn{2}{c}{\textbf{PointNet~\cite{qi2017pointnet}}} & \multicolumn{2}{c}{\textbf{w/o Two-step}} & \multicolumn{2}{c}{\textbf{Ours}} \\
& RMSE(T) & RMSE(R) & RMSE(T) & RMSE(R) & RMSE(T) & RMSE(R) & RMSE(T) & RMSE(R) & RMSE(T) $\downarrow$& RMSE(R)$\downarrow$ \\
\midrule
\textbf{Lid Covering}  & 0.098 & 18.23 & 0.245 & 70.06 & 0.234 & 65.32 & 0.117 & 18.74 & \textbf{0.090} & \textbf{16.12} \\
\textbf{Inserting}    & 0.157 & 41.22 & 0.234 & 62.25 & 0.234 & 66.80 & 0.164 & 41.11 & \textbf{0.142} & \textbf{38.03} \\
\textbf{Precision Placing} & 0.121 & 48.01 & 0.245 & 65.19  & 0.211 & 72.47  & 0.137 & 46.38  & \textbf{0.115} & \textbf{44.84} \\
\midrule
\textbf{ALL}            & 0.123 & 44.67 & 0.277 & 72.46  & 0.264 & 75.38  & 0.139 & 45.20  & \textbf{0.110} & \textbf{41.44} \\
\bottomrule
\end{tabular}%
}
\caption{\textbf{Ablation Study Results.} We compare various encoders including Vector Neuron DGCNN~\cite{deng2021vector}, DGCNN~\cite{wang2019dynamic}, PointNet~\cite{qi2017pointnet}, and our proposed two-scale Vector Neuron DGCNN. We also compare end-to-end networks with two-step networks to demonstrate the effectiveness of each component in our network design.}
\label{tab:exp ablation}
\end{table*}
\section{Ablation Study}
\label{sec:ablation}
In this section, we conduct comprehensive experiments to demonstrate the rationality of our network design and the effectiveness of each module.

\textbf{Encoder.} To validate the effectiveness of two-scale Vector Neuron(VN) DGCNN, we compare it with other encoders: VN DGCNN~\cite{deng2021vector}, DGCNN~\cite{wang2019dynamic}, and PointNet~\cite{qi2017pointnet}.

\textbf{Two-step network design.} We compare our method with an end-to-end approach, which jointly predicts the pose of $\mathcal{P_A}$ and $\mathcal{P_B}$. Specifically, we utilize $\mathcal{B_A}$ and the input point cloud of $\mathcal{P_A}$ and $\mathcal{P_B}$ to get the 6D pose of $P_A$, while using the same encoder to extract $\mathcal{E_B}$ and pass it through the same pose prediction head to predict the pose of $P_B$. 

As shown in in Table~\ref{tab:exp ablation}, we show the results of ablation studies on \textit{Lid covering}, \textit{Inserting} and \textit{Precision Placing}, which are more comprehensive and require cross-task generalization abilities. The performance declines in both translation and rotation when we removing our two-scale VN DGCNN encoder and change it to Vector Neuron DGCNN~\cite{deng2021vector}, DGCNN~\cite{wang2019dynamic}, Pointnet~\cite{qi2017pointnet}. It demonstrates that our encoder effectively exploits the advantage of SE(3) equivariance, enabling greater sample efficiency and more robust generalization abilities. Compared with version in Figure~\ref{fig:pipeline}, the experiment performance declines when we change our two-step network in a joint-learning manner, proving that our two-step network design can reduce error caused by jointly predictions and thereby is more effective.

\vspace{-0.2cm}
\section{Conclusion}
\label{sec:conclusion}

2BY2 is a significant step in bridging the gap between geometry-based assembly tasks and everyday object assemblies. With pose and symmetry annotations for 517 object pairs across 18 fine-grained tasks, 2BY2 sets a new benchmark for 3D assembly challenges. Our two-step pairwise SE(3) pose estimation framework, which leverages equivariant features, demonstrates superior performance over existing approaches in reducing both translation and rotation errors. Robot experiments further validate the method’s generalizability in practical 3D assembly scenarios. In conclusion, 2BY2 provides both a comprehensive benchmark and an effective framework, with the aim of inspiring and supporting more generalizable solution in robot manipulation.

{\small
\bibliographystyle{ieeenat_fullname}
\bibliography{l1_reference}
}
\clearpage \appendix \section{Appendix Section}

\subsection{2BY2 Dataset}

Unlike previous datasets like Breaking Bad and Neural Shape Mating ~\cite{sellan2022breaking, chen2022neural} which focus on assembly of object fragments, our ~\textit{\textbf{2BY2}} dataset focuses on pairwise assembly of daily objects with geometry and task variety, includes tasks that can be quite challenging for robot manipulation. For example ~\textit{Plug}, ~\textit{Bread}, ~\textit{flower} are very challenging in real world becuase they require precise pose alignment to achieve assembly success. 

In previous datasets such as Breaking Bad, the pose of each fragment depends on all the other fragments. However, in daily pairwise assembly task, the pose of the \textit{Object B}, such as bottle and toaster, is not affected by \textit{Object A}, such as cap and bread, and is only determined by the canonical space. In contrast, the pose of \textit{Object A} is influenced by the geometry and pose of \textit{Object B}. For instance, the pose of a cap is determined by the rim of the cup, while the pose of a piece of bread is dictated by the slot of the toaster. Consequently, previous methods that jointly predict the poses of two objects are not well-suited for daily pairwise assembly tasks. To address this, we propose a two-step paired network architecture that sequentially predicts the pose of each object, effectively mitigating pose errors introduced by joint pose prediction in prior approaches.

\subsubsection{Dataset Collection}

We segment, integrate, and pair meshes obtained online, scaling them to a global scale of \textit{3.0}. Each mesh pair is categorized into ~\textit{Object B} and ~\textit{Object A}, where ~\textit{Object B} serves as the receiving component, and ~\textit{Object A} functions as the fitting component. Similar to Breaking Bad~\cite{sellan2022breaking}, we triangulate each mesh using blender~\cite{blender} and use blue noise sampling method to extract the point cloud from the surface of each mesh, and use padding to make sure each dimension aligns with $(1024,3)$.

\subsubsection{Symmetry Annotation}

Each object is associated with a JSON file specifying its symmetry type. In this work, we account for two types of symmetry: axis symmetry along the \textit{x, y, z} axes, and rotational symmetry around the \textit{x, y, z} axes.

\subsubsection{Task Definition}

In the ~\textit{Lid Covering} category, ~\textit{Object A} refers to the lid, and ~\textit{Object B} refers to the corresponding body, including ~\textit{Kitchen}, ~\textit{Bottle}, ~\textit{Kettle}, ~\textit{Coffeemachine}, and ~\textit{Cup}.

In the \textit{Inserting} category:

\begin{itemize}
    \item In \textit{Plug}, \textit{Object A} is the plug, and \textit{Object B} is the socket.
    \item In \textit{Children's Toy}, \textit{Object A} is the block, such as cylinder and cone, and \textit{Object B} is the board with slots.
    \item In \textit{Letter}, \textit{Object A} is the mail, and \textit{Object B} is the postbox.
    \item In \textit{Bread}, \textit{Object A} is the bread, and \textit{Object B} is the toaster.
    \item In \textit{Nut}, \textit{Object A} is the bolt, and \textit{Object B} is the nut.
    \item In \textit{Coin}, \textit{Object A} is the coin, and \textit{Object B} is the piggy bank.
    \item In \textit{Key}, \textit{Object A} is the key, and \textit{Object B} is the lock.
    \item In \textit{USB}, \textit{Object A} is the cap, and \textit{Object B} is the USB body.
\end{itemize}

In the \textit{High Precision Placing} category:

\begin{itemize}
    \item In the \textit{Box} task, \textit{Object A} refers to the shoes, and \textit{Object B} refers to the box. The goal is to neatly place the shoes in the shoebox.
    \item In the \textit{Tissue} task, \textit{Object A} refers to the tissue, and \textit{Object B} refers to the tissue rack. The goal is to place the tissue on the rack.
    \item In the \textit{Flower} task, \textit{Object A} refers to the flower, and \textit{Object B} refers to the vase.
    \item In the \textit{Teapot} task, \textit{Object A} refers to the teapot, and \textit{Object B} refers to the tea tray. The goal is to neatly place the teapot on the tray.
    \item In the \textit{Position} task, \textit{Object A} refers to the cup, and \textit{Object B} refers to the coffee machine. The goal is to place the cup underneath the spout of the coffee machine.
\end{itemize}

\subsubsection{Definition of Canonical Pose in Different Tasks}

In all tasks except for \textit{Plug}, the canonical pose refers to the assembled state where the two objects are placed on the \textit{XY} plane under the influence of gravity, ensuring stable contact with the plane. Additionally, the positive Z-axis passes through the geometric center of the object's base, ensuring proper central and vertical alignment, as shown in Figure~\ref{fig:appendix_canonical}.

\begin{figure}[tp]
    \centering
    \includegraphics[width=\linewidth]{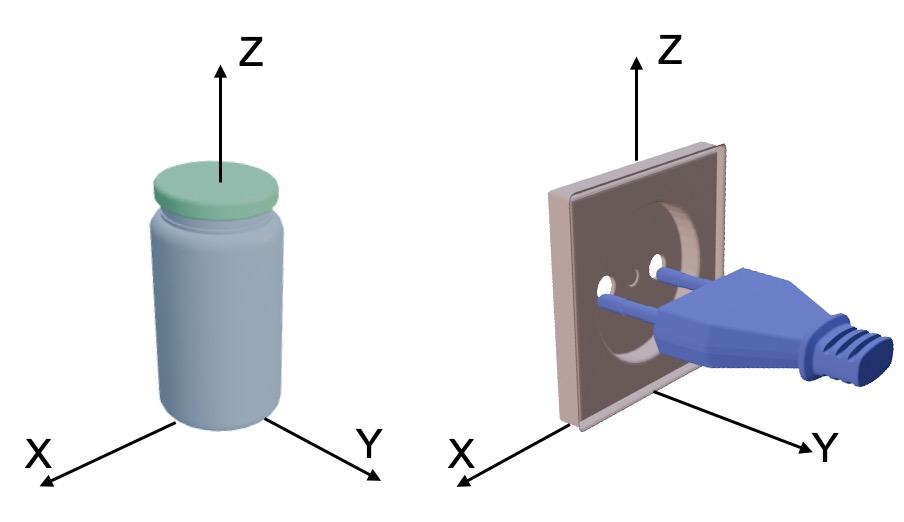}
    \caption{\textbf{The Definition of Canonical Pose.} The left image illustrates the canonical pose of the task \textit{bottle}, while the right image represents the canonical pose of \textit{plug}.}
    \label{fig:appendix_canonical}
\end{figure}

In the \textit{Plug} task, the canonical pose is defined as the state where the socket is placed on the \textit{XZ} plane, representing the wall, as shown in Figure \ref{fig:appendix_canonical}. 

Notably, in tasks where only a single relative pose is required—such as plugging into a socket which is fixed on the wall—the plug's pose can be determined through coordinate transformation, as illustrated in Section \ref{sec: real-robot experiment}.

\subsubsection{Data Splition}

\begin{figure}[tp]
    \centering
    \includegraphics[width=\linewidth]{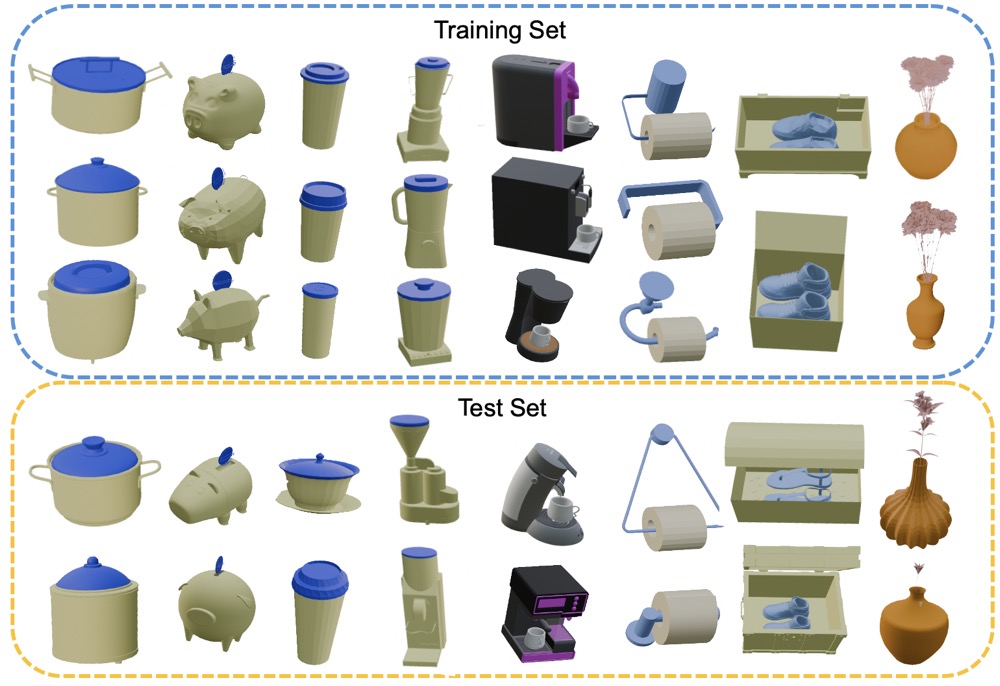}
    \caption{\textbf{Task Diversity Visualization.} From left to right, each column shows selected meshes from training set and test set of \textit{Kitchenport}, \textit{Coin}, \textit{Cup}, \textit{Coffeemachine}, \textit{Position}, \textit{Toilet}, \textit{Shoes}, \textit{Flower}.}
    \label{fig:supp_task_diversity}
\end{figure}

As described in the main paper, our \textbf{\textit{2BY2}} dataset includes \textit{18} fine-grained tasks, such as \textit{Bottle} and \textit{Children's Toy}, and \textit{4} tasks which require cross-category generalization ability, which is \textit{Lid Covering}, \textit{Inserting}, \textit{High Precision Placing} and \textit{All}. We ensure geometric diversity when assigning each object exclusively to either the training or test set, as shown in Figure \ref{fig:supp_task_diversity}.

For cross-category tasks like \textit{Lid Covering}, the training and test sets both include objects from its own categories, such as \textit{Kitchen}, \textit{Bottle}, \textit{Kettle}, \textit{Coffeemachine}, and \textit{Cup}. Similar applies to the \textit{Inserting} and \textit{High Precision Placing} tasks. For the \textit{All} task, both the training and test sets include all \textit{18} fine-grained tasks. 

For each of the \textit{18} fine-grained task, we maintain a training-to-test set ratio of approximately \textit{3:2}. For \textit{Lid Covering}, \textit{Inserting}, \textit{High Precision Placing} and \textit{All}, the ratio is controlled at roughly \textit{5:2}.

\subsection{Methodology}

\subsubsection{SE(3) Equivariant and SO(3) Invariant Feature}

Robots operate within a three-dimensional Euclidean space, where manipulation tasks inherently encompass geometric symmetries such as rotations. Recent works~\cite{huang2024imagination, xu2023unsupervised, zhu2023robot, wang2024equivariant, xue2023useek, gao2024riemann} leverage symmetry to enable robust learning and generalization. As illustrated in the main paper, SE(3) equivariant feature, which is extracted by our designed encoder, leverage symmetry to improve sample efficiency. In both branch, SE(3) equivariant features of $\mathcal{O}_B$ and $\mathcal{O}_A$ are used for object pose estimation.

SO(3) invariant features encode geometric shape information in the latent space, independent of the input point cloud’s orientation. In $\mathcal{B}_A$, the SO(3) invariant feature of $\mathcal{P}_B$ is extracted to facilitate the pose estimation of $\mathcal{P}_A$. Intuitively, the predicted pose of the bread is determined by the geometry of the toaster slot.

\subsection{Experiment}

\subsubsection{Data Augmentation}
During training, we apply SO(3) data augmentation to all methods, including both our approach and the baselines, which provides sufficient data for network convergence and ensures fair comparison. Notably, as pointed out by ~\cite{wang2022surprising}, although our network exhibits SE(3) equivariance, SO(3) data augmentation still benefits the learning process. 

\subsubsection{2BY2 Dataset Experiment}
\label{sec:supp exp}

Similar to Breaking Bad~\cite{sellan2022breaking}, we also use Chamfer Distance (CD) as our additional evaluation metric to validate the effectiveness our multi-step pairwise network.

\textbf{Evaluation Metric.} Chamfer Distance (CD)~\cite{chamferdistance} is a common metric used to measure the similarity between two point clouds or sets. It is widely applied in computer vision, 3D shape matching, point cloud alignment. More specifically, given two point clouds $P = \{p_1, p_2, \dots, p_m\}$ and $Q = \{q_1, q_2, \dots, q_n\}$, Chamfer Distance between $P$ and $Q$ is defined as:

\begin{equation}
CD(P, Q) = \frac{1}{|P|} \sum_{p \in P} \min_{q \in Q} \|p - q\|_2^2 + \frac{1}{|Q|} \sum_{q \in Q} \min_{p \in P} \|q - p\|_2^2
\end{equation}

More specifically, we use the average Chamfer Distance between the predicted $P'_B$ and ground truth $P_B$, and the predicted $P'_A$ and ground truth $P_A$:

\begin{equation}
CD = \frac{1}{2}(CD(P'_B, P_B) + CD(P'_A, P_A))
\end{equation}

\textbf{Results and Analysis.} As detailed in the main paper, we compare our multi-step pairwise network with SE-3 assembly~\cite{wu2023leveraging}, Puzzlefusion++~\cite{wang2024puzzlefusion++}, Jigsaw~\cite{lu2024jigsaw} and Neural Shape Mating~\cite{chen2022neural}. As shown in Table \ref{tab:exp_chamfer_distance} and Figure , our method consistently outperforms all baselines across 18 fine-grained tasks, demonstrating significantly improved alignment and geometric matching accuracy. This highlights the superior precision and effectiveness of our multi-step pairwise network. Moreover, in tasks such as \textit{Lid Covering}, \textit{Inserting}, \textit{Precision Placing}, and the overall \textit{All} category, our method achieves a substantial margin of improvement over the baselines, further indicating its robust generalization ability.

\begin{figure*}[tp]
    \centering
    \includegraphics[width=\linewidth]{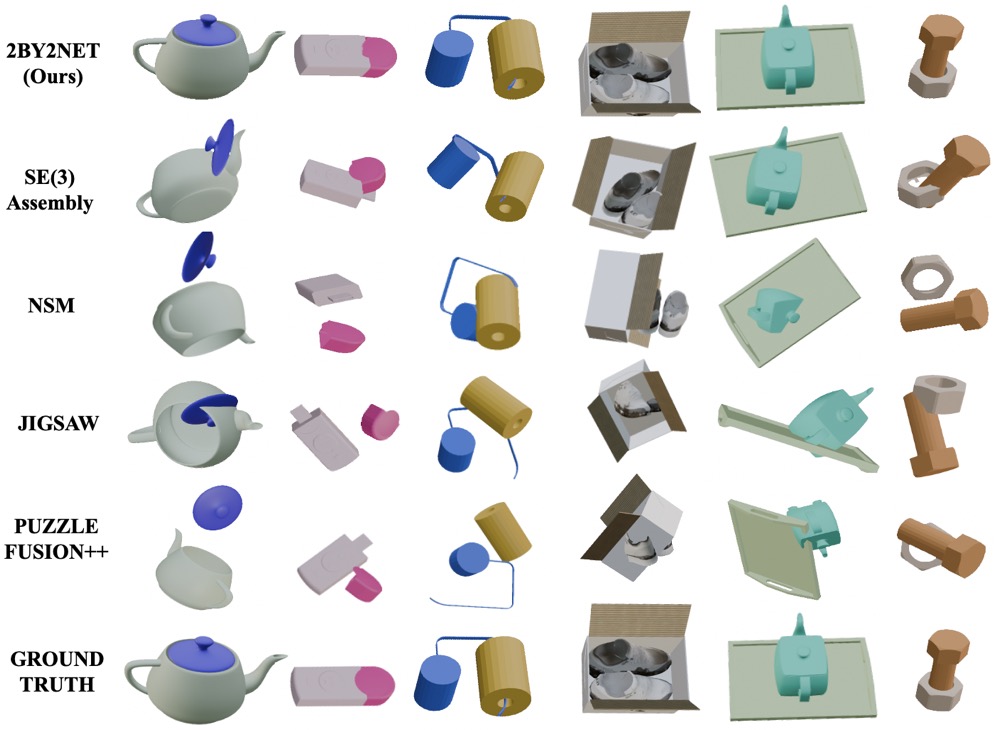}
    \caption{\textbf{Qualitative Results Comparison.} We highlight ~\textit{Kettle}, ~\textit{USB}, ~\textit{Toilet}, ~\textit{Shoes}, ~\textit{Teapot}, ~\textit{Nut} tasks to demonstrate our improved translation and rotation predictions compared to baseline methods.}
    \label{fig:appendix_canonical}
\end{figure*}

\begin{table*}[h]
\centering
\resizebox{1.0\textwidth}{!}{%
\begin{tabular}{l|c|c|c|c|c}
\toprule
\textbf{Task} & \textbf{Jigsaw~\cite{lu2024jigsaw}} & \textbf{Puzzlefusion++~\cite{wang2024puzzlefusion++}} & \textbf{NSM~\cite{chen2022neural}} & \textbf{SE(3)-Assembly~\cite{wu2023leveraging}} & \textbf{Ours} \\
 & CD & CD & CD & CD & CD $\downarrow$ \\
\midrule
\textbf{Lid Covering}  & 1.665 & 1.809 & 1.082 & 0.453 & \textbf{0.362} \\
Kitchenport  & 1.100 & 1.169 & 0.772 & 0.323 & \textbf{0.230} \\
Bottle       & 1.640 & 1.738 & 1.194 & 0.601  & \textbf{0.321} \\
Kettle        & 1.277 & 1.425 & 0.903 & 0.428 & \textbf{0.163} \\
Coffeemachine   & 1.290 & 1.394 & 1.178 & 0.394  & \textbf{0.189} \\
Cup    & 1.336 & 1.260 & 1.093 & 0.493 & \textbf{0.268}  \\
\textbf{Inserting}    & 0.712 & 0.842 & 0.860 & 0.431 & \textbf{0.278} \\
Plug               & 0.752 & 0.746 & 0.411 & 0.194  & \textbf{0.085} \\
Childrentoy                 & 1.037 & 0.917 & 0.874 & 0.814  & \textbf{0.791} \\
Letter         & 1.296 & 0.862 & 0.341 & 0.191  & \textbf{0.140} \\
Bread & 0.406 & 0.301 & 0.139 & 0.144  & \textbf{0.105} \\
Nut & 0.131 & 0.665 & 0.946 & 0.368  & \textbf{0.059} \\
Coin & 0.946 & 0.921 & 0.756 & 0.146  & \textbf{0.134} \\
Key & 0.603 & 0.829 & 0.441 & 0.149  & \textbf{0.032} \\
Usb & 0.541 & 0.656 & 0.508 & 0.327  & \textbf{0.266} \\
\textbf{Precision Placing} & 0.888 & 0.472 & 0.366 & 0.306  & \textbf{0.255} \\
Box  & 0.263 & 0.234 & 0.205  & 0.102 & \textbf{0.093}  \\
Tissue & 0.462 & 0.644 & 0.335 & 0.349  & \textbf{0.232} \\
Flower & 0.463 & 0.361 & 0.371 & 0.376  & \textbf{0.295} \\
Teaport & 0.577 & 0.475 & 0.345 & 0.157  & \textbf{0.069} \\
Position & 0.759 & 0.735 & 0.585 & 0.548  & \textbf{0.302} \\
\midrule
\textbf{ALL}            & 1.223 & 1.469 & 1.100 & 0.679  & \textbf{0.268} \\
\bottomrule
\end{tabular}%
}
\caption{\textbf{Quantitative Evaluation on \textit{2BY2} for Pairwise Object Assembly.} Our method outperforms the baseline across all 18 fine-grained assembly tasks, as well as demonstrating significant improvement on \textit{4} cross-category assembly tasks, including \textit{Lid covering}, \textit{Inserting}, \textit{Precision Placing} and \textit{All}. It achieves an average reduction of \textit{0.138} in Chamfer Distance.}
\label{tab:exp_chamfer_distance}
\end{table*}

\subsubsection{Real-robot Experiment}
\label{sec: real-robot experiment}
In some tasks in the real world, instead of two poses, only one relative pose is needed to solve the pairwise assembly task. For example, when plugging into the socket that is fixed to the wall, only the pose of the plug is needed. To resolve tasks like these, we first infer the socket's pose in our defined world frame. In this step, we are not rotating socket arbitrarily. Then estimate the plug’s target pose in defined world frame. The plug's target pose in the real world can be calculated using a coordinate transformation. 

Moreover, rather than relying on pre-defined grasping poses, numerous existing grasping methods, such as ~\cite{fang2023anygrasp, huang2023edge}, can generate adaptable grasps efficiently. The motion trajectory can then be computed using motion planning library.

\subsection{Ablation Study}

As detailed in the main paper, we compare our method on \textit{Lid covering}, \textit{Inserting}, and \textit{High precision placing} and \textit{All} task in \textbf{\textit{2BY2}} dataset with other encoders: Vector Neuron DGCNN~\cite{deng2021vector}, DGCNN~\cite{wang2019dynamic}, PointNet~\cite{qi2017pointnet} and an end-to-end approach which jointly predicts the pose of $P_A$ and $P_B$.

\textbf{Evaluation Metric.} Similar to Section \ref{sec:supp exp}, We choose Chamfer Distance (CD) as our additional evalution metric.

\textbf{Results and Analysis.} As shown in Table \ref{tab:supp ablation}, replacing our multi-scale VN DGCNN encoder with Vector Neuron DGCNN~\cite{deng2021vector}, DGCNN~\cite{wang2019dynamic}, or PointNet~\cite{qi2017pointnet} results in a performance drop, highlighting that our encoder better captures geometric features and exhibits greater sensitivity to pose transformations. Additionally, substituting our multi-step network with a joint-learning approach leads to an increase in Chamfer Distance, underscoring the effectiveness of our multi-step network design.

\begin{table*}[h!]
\centering
\resizebox{1.0\textwidth}{!}{%
\begin{tabular}{l|c|c|c|c|c}
\toprule
\textbf{Task} & \textbf{Vector Neuron DGCNN~\cite{deng2021vector}} & \textbf{DGCNN~\cite{wang2019dynamic}} & \textbf{PointNet~\cite{qi2017pointnet}} & \textbf{w/o Multi-step} & \textbf{Ours} \\
& Chamfer Distance & Chamfer Distance & Chamfer Distance & Chamfer Distance & Chamfer Distance $\downarrow$ \\
\midrule
\textbf{Lid Covering}  & 0.387 & 0.873 & 0.875 & 0.439 & \textbf{0.362}\\
\textbf{Inserting}    & 0.297 & 0.483 & 0.489 & 0.290 & \textbf{0.278} \\
\textbf{Precision Placing} & 0.274 & 0.864 & 0.729 & 0.283  & \textbf{0.255} \\
\midrule
\textbf{ALL}            & 0.294 & 0.806 & 0.816 & 0.307  & \textbf{0.268} \\
\bottomrule
\end{tabular}%
}
\caption{\textbf{Ablation Study Results.} We compare various encoders including Vector Neuron DGCNN~\cite{deng2021vector}, DGCNN~\cite{wang2019dynamic}, PointNet~\cite{qi2017pointnet}, and our proposed multi-scale Vector Neuron DGCNN. We also compare end-to-end networks with multi-step networks to demonstrate the effectiveness of each component in our network design.}
\label{tab:supp ablation}
\end{table*}

\subsection{Limitations and Future Works}

The current design of our network is primarily constrained by the scope of the \textbf{\textit{2BY2}} dataset, which could be further expanded to include a wider range of tasks and more complex everyday scenarios. Additionally, rather than hardcoding the grasping pose, a policy network for robotic manipulation could be trained using the \textbf{\textit{2BY2}} dataset. Furthermore, the network architecture can be optimized to reduce computational overhead, improving its suitability for real-time robotic operations.

\end{document}